\documentclass[11pt]{article}

\usepackage[preprint]{acl}

\usepackage{times}
\usepackage{latexsym}

\usepackage[T1]{fontenc}
\usepackage[utf8]{inputenc}

\usepackage{microtype}

\usepackage{inconsolata}

\usepackage{graphicx}

\usepackage{amsmath}
\usepackage{amsfonts}
\usepackage{booktabs}
\usepackage{multirow}
\usepackage{makecell}

\title{Self-Ensembling Vision-Language Models for Chart Data Extraction}

\author{
  \textbf{Thomas Berkane}, \quad \textbf{Qianyi Wang}, \quad \textbf{Maimuna S. Majumder} \\
  Computational Health Informatics Program, Boston Children's Hospital \& Harvard Medical School \\
  \small{\textbf{Correspondence:} \href{mailto:thomas.berkane@childrens.harvard.edu}{thomas.berkane@childrens.harvard.edu}}
}

\begin{document}
\maketitle
\begin{abstract}
Charts effectively convey quantitative information, but the underlying data are often locked in image form, hindering reuse and analysis. Manually digitizing charts is time-consuming and error-prone, motivating automatic chart-to-table extraction. Recent approaches use specialized vision-language models (VLMs), yet performance still lags on charts with many datapoints or substantial stylistic variation. We propose a VLM self-ensembling method that repeatedly samples multiple tabular outputs from the same VLM for a fixed chart image and aggregates them at the level of individual table cells. We align candidate tables and take per-cell medians over numerical values to produce a more accurate consensus table. Our method also includes convergence detection to stop sampling once the aggregated table stabilizes, and uncertainty estimation based on dispersion across samples to help users assess extraction reliability. Because existing chart extraction benchmarks contain relatively simple plots with limited room for improvement, we introduce WB-ChartExtract, a new benchmark built from World Bank data with more complex and stylistically diverse charts; on average, its charts contain 7$\times$ more datapoints than those in the ChartQA benchmark. Across both ChartQA and WB-ChartExtract, our approach improves extraction accuracy over single-pass VLM outputs, yielding up to 23\% relative improvement on WB-ChartExtract after ensembling. More broadly, our method helps unlock tabular data previously siloed in chart images, enabling downstream analysis and reuse.
\end{abstract}

\section{Introduction}
\label{sec:introduction}
Charts and plots are a primary medium for communicating quantitative information in scientific articles, policy reports, news stories, and online dashboards. Unlike tables, charts typically appear only as rasterized images in PDFs or on the web, making the underlying numbers difficult to access: they cannot be searched, joined with other sources, or reused for downstream analysis. Recovering these data at scale is increasingly important for meta-analysis, model validation, and real-time monitoring, where source data are often missing or inaccessible.

Manual digitization with tools such as WebPlotDigitizer \cite{WebPlotDigitizer} is possible but slow and error-prone, especially for large corpora or visually complex figures. This has motivated automatic chart-to-table extraction, where a system reconstructs a structured table of datapoints from a chart image. Early systems \citep{early,revision} used engineered pipelines combining computer vision, optical character recognition (OCR), and heuristic reasoning about axes and legends. More recent work \citep{tinychart,onechart,chartvlm} leverages vision-language models (VLMs) that can directly ``read'' charts and emit a textual table representation. While VLM-based methods simplify the stack and have improved performance and generalization on several benchmarks \citep{chartqa,plotqa}, they still struggle with dense datapoints, multiple overlaid series, or idiosyncratic labeling.

These failure modes reflect a broader pattern in generative models: a single forward pass can be brittle \cite{selfconsistency,ttc}. When prompted multiple times on the same chart, a VLM can return noticeably different tables---dropping or adding points or entire series, or misreading values---yet these outputs are often partially correct and complementary. We verify this empirically on our benchmark: running a chart-specialized VLM (TinyChart) 20 times per chart on WB-ChartExtract, 99.4\% of charts produced at least one differing extraction, and 49.0\% had different table dimensions across runs. Existing chart extraction methods typically ignore this variability, relying on a single ``best guess,'' and they provide little signal about extraction reliability, which is problematic for realistic, cluttered figures.

We propose a complementary approach that embraces VLM stochasticity by ensembling outputs at the level of individual table cells. Our method is model-agnostic, can be layered on top of existing chart extraction systems, and applies to any chart type the underlying model can handle. Concretely, we repeatedly prompt a VLM on the same chart image, align candidate tables at the cell level, and compute per-cell medians over numerical values to form a consensus table. We also introduce a convergence detection mechanism to stop sampling once additional predictions are unlikely to change the ensemble, and an uncertainty estimate based on variability across samples to quantify extraction reliability.

Progress on chart extraction also depends on benchmarks with challenging, stylistically varied charts. Yet widely used datasets such as ChartQA offer increasingly limited room for improvement: state-of-the-art VLMs already score highly, in part because many plots are simple, with few datapoints, limited stylistic variation, and numbers printed directly on the chart. To provide a more demanding testbed, we introduce WB-ChartExtract, a benchmark of synthetic charts built from real World Bank data that contain on average 7$\times$ more datapoints than ChartQA charts, span four chart types and four rendering libraries, and include no printed value labels (\S\ref{sec:datasets}).

We evaluate our self-ensembling approach on both ChartQA and WB-ChartExtract. Across datasets and underlying models, ensembling consistently improves accuracy over single-pass VLM extraction, with gains of up to 23\% relative on WB-ChartExtract. We further analyze which error types our approach corrects and which remain challenging. Importantly, we show that convergence detection is well calibrated and supports a controllable trade-off between computational cost and marginal accuracy gains via practical early stopping. Finally, we demonstrate that our uncertainty estimate is empirically inversely correlated with extraction accuracy.

Our contributions are fourfold:
\begin{enumerate}
    \item A model-agnostic self-ensembling method for chart data extraction.
    \item A convergence detection mechanism that identifies when additional samples are unlikely to change the aggregated result.
    \item Ensemble uncertainty estimates to help users gauge extraction reliability.
    \item WB-ChartExtract, a new benchmark constructed from real-world data that is, to our knowledge, the most challenging to date.
\end{enumerate}

Together, these contributions move chart-to-table extraction closer to real-world deployment.

\section{Background and Related Work}
Chart data extraction (also called \emph{chart-to-table} or \emph{derendering}) aims to recover the numerical values in a chart image along with their semantic structure (e.g., axis values and series names). Early systems decomposed the task into multi-stage pipelines combining computer vision, OCR, and heuristics to map pixels to values \citep{early,revision,early2,early3}. While effective when figures match anticipated templates, these pipelines are often brittle under modest stylistic variation (e.g., layout, font, or color).

Recent work instead uses VLMs to generate a textual table representation directly from the chart image \citep{deplot,matcha,chartllama,chartast,onechart,tinychart,chartvlm}. However, VLMs often struggle with fine-grained visual distinctions \citep{amplified,spatialreasoninghard}, including in chart settings \citep{plottwist}. Consequently, one-pass generation remains error-prone on complex charts: models may omit datapoints, hallucinate values, misread axes, or inconsistently name series across runs. Our approach is designed to sit atop either specialized chart extraction models or generalist VLMs, repeatedly querying the underlying model and aggregating outputs to improve robustness.

\paragraph{Benchmarks.}
Common chart-extraction datasets include ChartQA \cite{chartqa}, PlotQA \cite{plotqa}, and ChartX \cite{chartvlm}. ChartQA contains real-world charts, but many instances are visually simple (e.g., values printed directly on bars). PlotQA and ChartX provide large-scale synthetic charts with broader chart-type coverage, but limited stylistic diversity and complexity. In contrast, WB-ChartExtract does not print values directly on marks, includes longer series and multiple series per chart, spans four chart types and four rendering libraries, and exhibits substantial stylistic variance.

\section{Methodology}
\begin{figure}[t]
    \centering
    \includegraphics[width=0.48\textwidth]{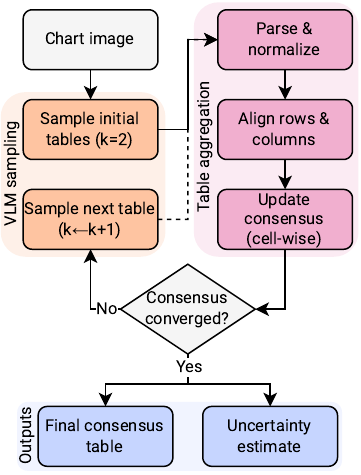}
    \caption{Iterative self-ensembling for chart-to-table extraction. We sample tables from a base VLM, parse/normalize and align rows/columns, and update a cell-wise consensus until convergence, producing a final table and uncertainty estimate based on the median absolute deviation (MAD).}
    \label{fig:method_overview}
\end{figure}

Given a chart image, our goal is to recover the underlying table of numeric datapoints. Our approach is \emph{model-agnostic}: we repeatedly prompt a base VLM to produce tabular outputs, then align and ensemble these outputs at the level of individual table cells to obtain a consensus table. Compared to single-pass extraction, self-ensembling suppresses outlier predictions (e.g., hallucinated or missing datapoints) and yields a natural uncertainty signal. Figure~\ref{fig:method_overview} provides an end-to-end overview.

\subsection{Base Model Sampling}
For each chart image, we query a base VLM (by default, Llama~4~Scout~\cite{llama}\footnote{We developed primarily with Llama~4~Scout due to its efficiency and low inference cost on Groq \cite{groq}, but the method is model-agnostic.}) with the following prompt:
\begin{quote}\small
Here is an image of a chart. 
Please extract the numerical data it represents and return it in TSV (tab-separated values) format with appropriate headers. 
Copy the headers exactly as they are in the image. 
IMPORTANT: For the TSV, use tab (\texttt{\textbackslash t}) as the separator.
Remember: The sole output should be the TSV table surrounded by ```tsv ```. Nothing else.
\end{quote}
We sample at temperature $T$ (a hyperparameter; Section~\ref{sampling-temperature}). For each image, we generate two initial tables, parse, align, and aggregate them (as described below). We continue sampling and updating the ensemble until either:
(i) for \emph{patience}$=2$ consecutive updates, at least \emph{coverage}$=95\%$ of aggregated cell values change by no more than a relative \emph{tolerance}$=1\%$ between successive ensemble updates (sensitivity analysis in Appendix~\ref{sec:sensitivity}); or
(ii) a maximum number of samples $K_{\max}$ is reached. We use $K_{\max}=20$ by default based on our computational budget, but this can be adjusted (Section~\ref{sec:maximum-ensemble-size}).

\subsection{Parsing and Normalization}
Each sample is a TSV-like text block and may contain formatting errors (e.g., inconsistent column counts). We first parse as TSV; if parsing fails due to ragged rows, we repair by padding or truncating rows to the modal column count.

After parsing, we treat (i) the first row as column headers, (ii) the first column as row headers (index), and (iii) remaining cells as numeric values. We convert value cells to numeric, stripping common artifacts (commas, currency symbols, percent signs); non-numeric strings (including empty cells) are treated as missing and set to \texttt{nan}. If needed, we transpose the table to enforce more rows than columns, which simplifies alignment for time-on-x charts with multiple series.

\subsection{Row and Column Alignment via Clustering}
Across samples, tables may differ in row/column order and labels may be noisy (e.g., spelling, abbreviations). We therefore align tables by clustering row and column labels (independently) across samples to form canonical row and column groups.

We measure label similarity with ANLS \citep[average normalized Levenshtein similarity]{anls}, yielding $\mathrm{sim}(\ell_1,\ell_2)\in[0,1]$ \citep{deplot,deplot_cited}. Clusters must satisfy: (i) grouped labels are similar under $\mathrm{sim}$, and (ii) no cluster contains two labels from the same sampled table (to avoid merging within-table duplicates).

We use greedy clustering with threshold $\tau=0.5$ (following \citet{deplot,deplot_cited}). Each cluster $c$ is represented by $\mathrm{rep}(c)$, the first label assigned to $c$:
\begin{enumerate}
    \item Initialize an empty set of clusters.
    \item For each label $\ell$, assign it to an existing cluster $c$ if (i) $\mathrm{sim}(\ell,\mathrm{rep}(c))\ge\tau$ and (ii) $c$ contains no label from the same sampled table as $\ell$. If multiple clusters qualify, choose the one maximizing $\mathrm{sim}(\ell,\mathrm{rep}(c))$.
    \item If no cluster qualifies, create a new cluster containing $\ell$.
\end{enumerate}

We then prune clusters that appear in fewer than 20\% of tables (default; ablation in Appendix~\ref{sec:cluster-pruning-threshold}) to remove spurious labels that would otherwise introduce extra rows/columns. For each remaining cluster, we define its canonical label as the most frequent label string in the cluster (ties broken at random) and use these canonical labels as the aggregated table’s row/column names.

\subsection{Cell-Wise Aggregation}
Let $\mathcal{R}=\{r_1,\dots,r_{|\mathcal{R}|}\}$ and $\mathcal{C}=\{c_1,\dots,c_{|\mathcal{C}|}\}$ be the retained row and column clusters. For each aligned cell $(r,c)\in\mathcal{R}\times\mathcal{C}$, we collect all numeric predictions from sampled tables whose row label falls in $r$ and column label falls in $c$. Let $V_{r,c}$ denote this set (ignoring \texttt{nan}). We aggregate as $\hat{y}_{r,c}=\mathrm{median}(V_{r,c})$ (ablation in Appendix~\ref{sec:aggregation-method}); if $V_{r,c}$ is empty, we output \texttt{nan}. The final table orders row and column clusters by lexicographic sort of their canonical labels.

\subsection{Uncertainty Estimation}
Self-ensembling yields a natural uncertainty signal from disagreement across samples: consistent values imply reliable medians, while high variability indicates brittle extraction and additional verification (or more samples) may be warranted.

For each aligned cell $(r,c)$, we compute uncertainty as the median absolute deviation (MAD) over $V_{r,c}$, normalized by the magnitude of the ensembled value:
\[
\begin{aligned}
u_{r,c}
&=
\frac{\mathrm{MAD}_{r,c}}{|\hat{y}_{r,c}|}, \\
\text{where}\quad
\mathrm{MAD}_{r,c}
&=
\mathrm{median}_{v\in V_{r,c}} \left|v-\hat{y}_{r,c}\right|.
\end{aligned}
\]
To avoid division by zero, we exclude cells with $\hat{y}_{r,c}=0$ when computing table-level uncertainty summaries. Relative MAD is scale-invariant, making uncertainty comparable across charts and across cells with different magnitudes.

To obtain a single uncertainty score per chart, we summarize $\{u_{r,c}\}$ over cells with non-missing predictions and $\hat{y}_{r,c}\neq 0$ using: $U_{\text{med}}$ (median), $U_{\text{mean}}$ (mean), and $U_{\text{max}}$ (maximum). These capture complementary failure modes: $U_{\text{med}}$ reflects typical uncertainty, $U_{\text{mean}}$ reflects overall dispersion, and $U_{\text{max}}$ flags catastrophic disagreement in any cell.

\section{Experiments}
\label{sec:experiments}
\subsection{Datasets}
\label{sec:datasets}
\paragraph{ChartQA.}
We evaluate on ChartQA, the most widely used benchmark for chart data extraction. Its test set contains 1,509 real-world chart images, but has notable limitations: charts are often visually simple, with relatively few datapoints, limited stylistic diversity, and values frequently printed directly on the chart, allowing models to rely on text rather than chart geometry and axes. As a result, the benchmark is somewhat saturated. ChartQA’s ground-truth tables are also often noisy, reducing evaluation reliability.

\begin{figure}[t]
    \centering
    \includegraphics[width=0.48\columnwidth]{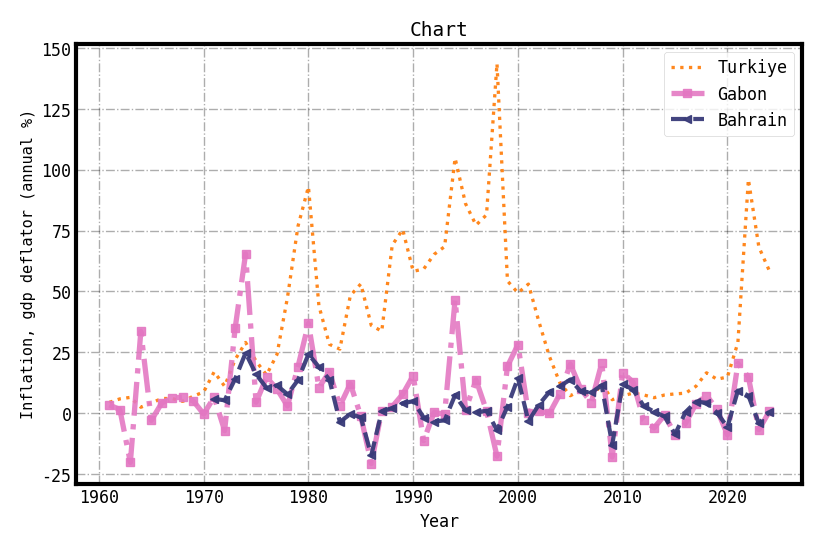}%
    \hfill
    \includegraphics[width=0.48\columnwidth]{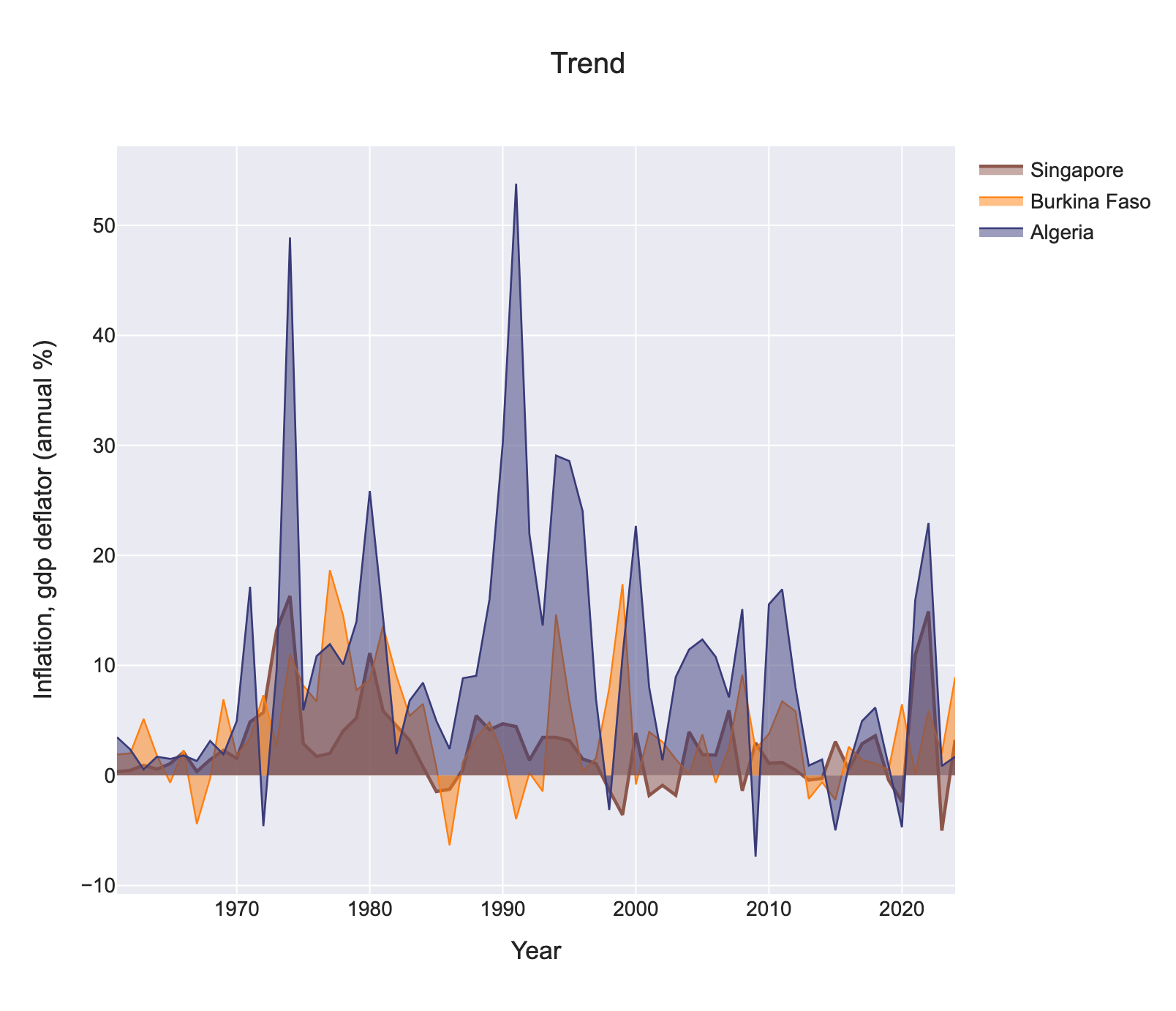}

    \vspace{0.5em}

    \includegraphics[width=0.48\columnwidth]{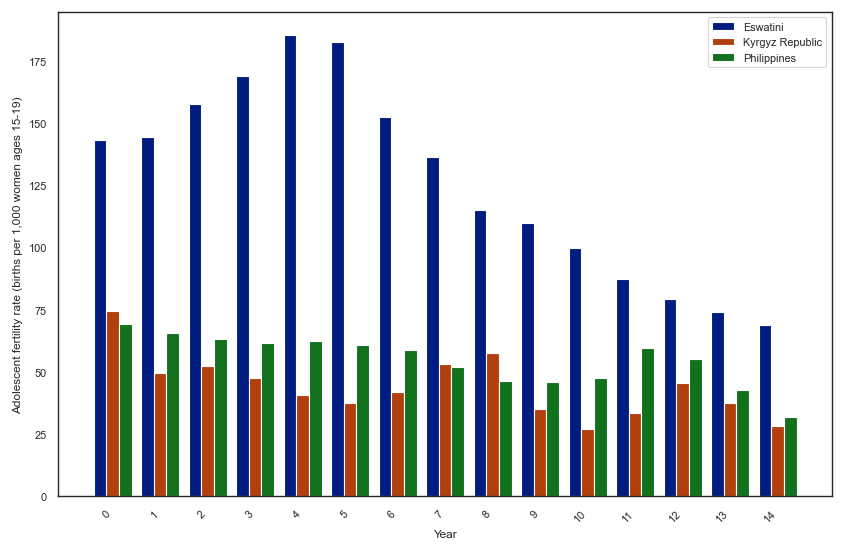}%
    \hfill
    \includegraphics[width=0.48\columnwidth]{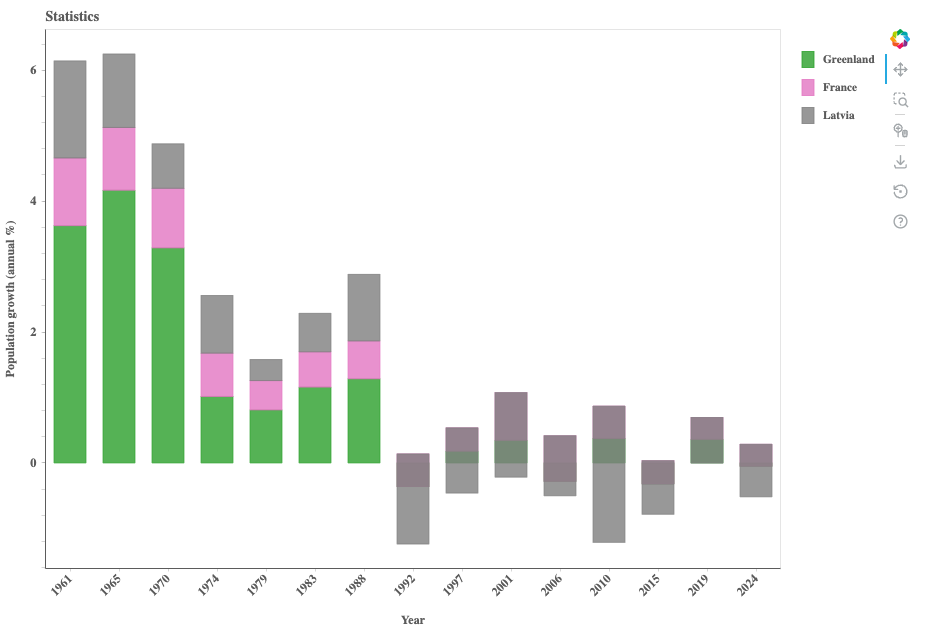}
    \caption{Examples of charts in WB-ChartExtract, showing four chart types rendered by four different libraries (clockwise from top-left: line/Matplotlib, area/Plotly, stacked bar/Bokeh, grouped bar/Seaborn).}
    \label{fig:WB-ChartExtract_examples}
\end{figure}

\paragraph{WB-ChartExtract.}
To address these limitations, we introduce WB-ChartExtract, a new benchmark constructed from World Bank Data Bank time series \cite{worldbank} spanning 52 indicators, 218 countries, and 65 years. We prune series with more than half missing values or where there are missing values in the interior (i.e., not just at the ends) of the series. We then create 1,000 datasets by repeatedly sampling one indicator and 2--3 countries at random and collecting the corresponding series, using each underlying series at most once. Each dataset is rendered as one of four chart types---line, area, grouped bar, or stacked bar---using one of four plotting libraries: Matplotlib \cite{matplotlib}, Seaborn, Plotly, or Bokeh. Chart type and library are assigned uniformly at random, yielding 62--63 charts per combination. Within each chart, we randomize font family and size, color palette, grid presence and style, line styles, markers, transparency, and figure size. This yields 1,000 charts with diverse visuals, chart types, and rendering styles, along with clean ground-truth tables. Its difficulty arises from several complementary factors rather than any single one: higher data density (on average 7$\times$ more datapoints than ChartQA, particularly for line and area charts), the absence of printed value labels (removing the OCR shortcut available on many ChartQA plots), multiple overlaid series, chart-type diversity requiring different visual inference strategies, and cross-library stylistic variation. Figure~\ref{fig:WB-ChartExtract_examples} shows example charts.

Synthetic generation is a deliberate choice: it yields clean ground-truth tables at scale, which is otherwise difficult to obtain for real-world charts since the underlying data is rarely available---reflected in the noisy ChartQA annotations noted above. Existing synthetic benchmarks do not fill this gap, as PlotQA \cite{plotqa} and FigureQA \cite{figureqa} target question answering rather than structured extraction, with fewer series, shorter series lengths, and fewer datapoints per chart.

To ensure full reproducibility, we release the complete benchmark---all chart images and clean ground-truth tables (under CC BY 4.0; World Bank data is CC BY 4.0)---together with our chart-generation, extraction, ensembling, and evaluation code and configuration files.\footnote{Code: \url{https://github.com/tberkane/vlm-ensemble-chart}. Data: \url{https://huggingface.co/datasets/tberkane/WB-ChartExtract}.}

\subsection{Evaluation Metric}
\label{sec:evaluation-metric}
We evaluate using \emph{Relative Mapping Similarity} (RMS$_{F1}$) \citep{deplot}, which treats each table as an unordered set of (row header, column header, value) triples and computes a minimum-cost bipartite matching between predicted and ground-truth entries. Per-entry similarity multiplies a header score (one minus thresholded normalized Levenshtein distance, threshold $\tau$) with a value score (one minus clipped relative error, threshold $\theta$). We report the harmonic mean of mapping-level precision and recall, set $\tau=0.5$ and $\theta=0.1$ following \citet{deplot}, and take the maximum over the predicted table and its transpose for transposition invariance. Full definition in Appendix~\ref{sec:rms-details}.

\subsection{Main Results}
We compare against single-pass predictions from three chart-specialized models (OneChart \citep{onechart}, TinyChart \citep{tinychart}, DePlot \citep{deplot}), three open-source VLMs (Qwen3-VL 235B A22B Instruct \citep{qwen}, Llama~4~Scout \citep{llama}, Seed~1.6~Flash \citep{seed}), and three closed-source VLMs (GPT-5.1 \citep{gpt}, Claude Opus~4.6 \citep{claude}, Gemini~2.5 Pro \citep{gemini}). To demonstrate model-agnosticism, we apply self-ensembling on top of TinyChart, DePlot, Qwen3-VL, Llama~4~Scout, and Seed~1.6~Flash, a spread that spans chart-specialized models and general-purpose open-source VLMs. The closed-source frontier models (GPT-5.1, Claude Opus~4.6, Gemini~2.5 Pro) serve as strong single-pass reference points that characterize the difficulty ceiling of each benchmark. For convergence detection, we use default parameters with a tolerance of $1\%$.

Table~\ref{tab:main_results} reports RMS$_{\text{F1}}$ on ChartQA and WB-ChartExtract. Single-pass performance on WB-ChartExtract spans a wide range (23.06--87.83), reflecting the benchmark's difficulty and diversity. Gemini~2.5 Pro is the strongest general-purpose model on WB-ChartExtract (TinyChart leads ChartQA at 95.20 thanks to in-domain fine-tuning, as discussed below; Qwen3-VL leads among general-purpose models on ChartQA at 91.43), achieving 87.83 on WB-ChartExtract, while weaker models show a large gap between the two.

The reversal between TinyChart and the general-purpose models across the two benchmarks reflects training-data alignment: TinyChart is fine-tuned on ChartQA's training split, inflating its in-distribution score, whereas the general-purpose models have no exposure to either benchmark. On WB-ChartExtract, where no model has seen the underlying data and the benchmark spans four chart types rendered by four libraries, the strongest general-purpose models' broader visual reasoning yields the best single-pass results. This further supports WB-ChartExtract as a complementary benchmark that probes general chart understanding rather than benchmark-specific fit.

Self-ensembling consistently improves over single-pass on WB-ChartExtract across all five base models: Qwen3-VL improves from 52.91 to 56.23 (+3.32), Seed~1.6~Flash from 35.08 to 43.17 (+8.09), Llama~4~Scout from 30.41 to 33.14 (+2.73), DePlot from 23.06 to 24.57 (+1.51), and TinyChart from 28.61 to 29.37 (+0.76). The relative gains span roughly 2.7\% (TinyChart) to 23.1\% (Seed~1.6~Flash), meaningful improvements given the benchmark's difficulty. On ChartQA, self-ensembling improves the three general-purpose VLMs (Seed~1.6~Flash 81.62$\rightarrow$87.04, Llama~4~Scout 75.13$\rightarrow$77.10, Qwen3-VL 91.43$\rightarrow$93.21) as well as the DePlot specialist (88.32$\rightarrow$89.02), while the gain is essentially zero for TinyChart (95.20$\rightarrow$95.28, +0.08), which is already saturated on its in-domain training distribution and leaves no headroom for ensembling to recover. These improvements are corroborated by two additional value-centric metrics, RNSS and RD, reported in Appendix~\ref{sec:rnss-rd}.

\begin{table}[t]
    \centering
    \small
    \setlength{\tabcolsep}{7pt}
    \renewcommand{\arraystretch}{1.12}
    \begin{tabular}{lcc}
        \toprule
        \textbf{Method} & \textbf{ChartQA} & \textbf{WB-ChartExtract} \\
        \midrule
        OneChart & 35.93 & 26.49 \\
        \addlinespace
        DePlot & 88.32 & 23.06 \\ 
        \quad + Self-ens. (ours) & 89.02 & 24.57 \\ 
        \addlinespace
        TinyChart & 95.20 & 28.61 \\
        \quad + Self-ens. (ours) & 95.28 & 29.37 \\
        \addlinespace
        Qwen3-VL & 91.43 & 52.91 \\
        \quad + Self-ens. (ours) & 93.21 & 56.23 \\
        \addlinespace
        Seed~1.6~Flash & 81.62 & 35.08 \\
        \quad + Self-ens. (ours) & 87.04 & 43.17 \\
        \addlinespace
        Llama~4~Scout & 75.13 & 30.41 \\
        \quad + Self-ens. (ours) & 77.10 & 33.14 \\
        \addlinespace
        GPT-5.1 & 84.80 & 51.26 \\
        Claude Opus 4.6 & 87.71 & 60.99 \\
        Gemini 2.5 Pro & 88.23 & 87.83 \\
        \bottomrule
    \end{tabular}
    \caption{Main results (RMS$_{\text{F1}}$; higher is better). ``Self-ens.'' denotes our self-ensembling procedure applied on top of a base model's direct predictions.}
    \label{tab:main_results}
\end{table}

\paragraph{Cost--accuracy tradeoff.}
Table~\ref{tab:cost} reports API cost per benchmark. Self-ensembling increases cost by roughly 4--16$\times$ depending on the model and dataset, but absolute costs remain modest---under \$15 for an entire benchmark in every ensembled configuration. The convergence detection mechanism (\S\ref{sec:convergence-detection-ablation}) further mitigates this by stopping early for easier samples. Appendix~\ref{sec:cost-figure} (Figure~\ref{fig:cost_efficiency}) visualizes this accuracy--cost tradeoff, showing that self-ensembling shifts every base model up and to the right, trading a higher per-image cost for consistent accuracy gains.

\begin{table}[t]
    \centering
    \small
    \setlength{\tabcolsep}{5pt}
    \renewcommand{\arraystretch}{1.05}
    \begin{tabular}{lcc}
        \toprule
        \textbf{Method} & \textbf{ChartQA} & \textbf{WB-ChartExtract} \\
        \midrule
        OneChart & \$0.00 & \$0.00 \\
        TinyChart & \$0.00 & \$0.00 \\
        \quad + Self-ens. & \$0.00 & \$0.00 \\
        DePlot & \$0.00 & \$0.00 \\
        \quad + Self-ens. & \$0.00 & \$0.00 \\
        \addlinespace
        Qwen3-VL & \$0.42 & \$0.63 \\
        \quad + Self-ens. & \$1.78 & \$9.03 \\
        \addlinespace
        Seed~1.6~Flash & \$0.76 & \$1.06 \\
        \quad + Self-ens. & \$5.30 & \$14.63 \\
        \addlinespace
        Llama~4~Scout & \$0.24 & \$0.28 \\
        \quad + Self-ens. & \$2.29 & \$4.47 \\
        \addlinespace
        GPT-5.1 & \$2.93 & \$4.55 \\
        Claude Opus 4.6 & \$32.13 & \$39.17 \\
        Gemini 2.5 Pro & \$7.04 & \$11.09 \\
        \bottomrule
    \end{tabular}
    \caption{Total API inference cost (USD) per benchmark.}
    \label{tab:cost}
\end{table}

\subsection{Error Analysis}
\label{sec:error_analysis}

\begin{figure}[t]
    \centering
    \includegraphics[width=0.48\textwidth]{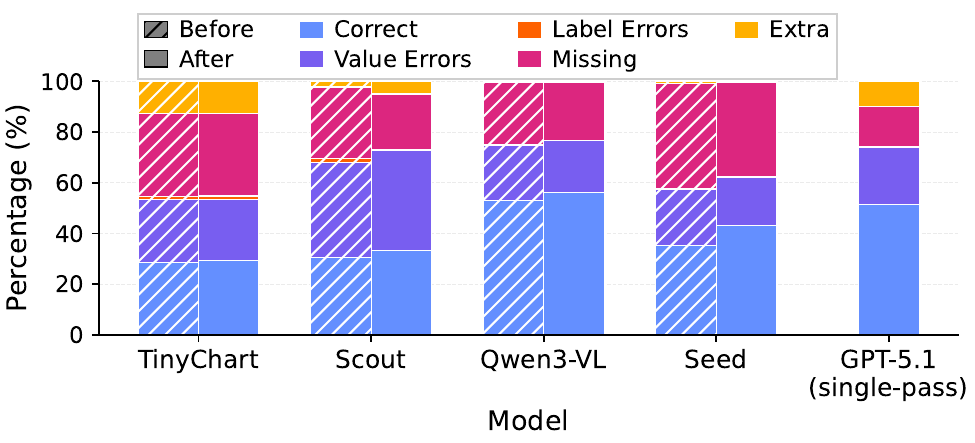}
    \caption{Error-type breakdown on WB-ChartExtract for each base model before vs.\ after self-ensembling.}
    \label{fig:error_breakdown}
\end{figure}

To understand how self-ensembling improves RMS$_{\text{F1}}$, we decompose each model’s output into \textbf{Correct} (example-level F1) plus four error categories that together sum to 100\%: \textbf{Value Errors} (numeric mismatch under the relative error distance), \textbf{Label Errors} (imperfect label matching under ANLS), \textbf{Missing} (ground-truth datapoints with no matched prediction), and \textbf{Extra} (predicted datapoints with no matched ground-truth). Figure~\ref{fig:error_breakdown} shows the average breakdown before and after ensembling on WB-ChartExtract for TinyChart, Llama~4~Scout, Qwen3-VL, and Seed~1.6~Flash, together with the single-pass breakdown for GPT-5.1.

\textbf{Label Errors} are negligible ($\le$1.4\%), and errors are dominated by \textbf{Value Errors} (19--40\%) and \textbf{Missing} (22--42\%). Ensembling improves \textbf{Correct} for all four base models shown (by 0.8--8.1\,pp), but the mechanism varies by model. For Qwen3-VL and Seed~1.6~Flash, gains come primarily from reduced \textbf{Value Errors} (1.6--3.1\,pp) together with fewer \textbf{Missing} datapoints (1.5--4.5\,pp). For Llama~4~Scout, the gain is instead driven by recovering \textbf{Missing} structure (28.3$\rightarrow$21.8\%), at the cost of a slight rise in \textbf{Value Errors} and \textbf{Extra} datapoints as more candidate rows are retained. \textbf{Extra} stays small for Qwen3-VL and Seed~1.6~Flash ($<$1\%), is moderate for Scout ($\approx$5\%), and is substantially larger for TinyChart ($\approx$13\%), reflecting frequent spurious predicted datapoints when the chart-specialized model is applied to out-of-distribution charts. Even a strong frontier model like GPT-5.1 is bottlenecked by \textbf{Value Errors} (22.8\%) and tends to over-generate datapoints (\textbf{Extra} 10.1\%). Across both weak and strong models, then, the dominant failure mode is precise numeric reading rather than chart structure---especially on area and line charts, where values must be inferred from curves rather than discrete bar heights.

\paragraph{Breakdown by chart type.}
Table~\ref{tab:wb_by_type} reports single-pass RMS$_{\text{F1}}$ on WB-ChartExtract by chart type (known by construction; see Section~\ref{sec:datasets}). Bar charts are the easiest categories and area charts the hardest. Grouped bar charts have the highest scores across all models (e.g., 95.18 for Gemini), likely because the subsampled year axis yields fewer datapoints with discrete, easy-to-read bar heights. Area charts are hardest (e.g., 79.20 for Gemini, 13.94 for Llama~4~Scout), as overlapping or stacked filled regions make it difficult to decompose individual series values. Weaker models show a particularly steep drop from bar to area charts: Llama~4~Scout falls from 48.39 on grouped bars to 13.94 on area, while Gemini's drop is more modest (95.18 to 79.20). This suggests that chart-type diversity in WB-ChartExtract exposes genuine differences in visual reasoning ability that a single chart type would miss. The analogous breakdown on ChartQA is in Appendix~\ref{sec:chartqa-by-type}.

\begin{table}[t]
    \centering
    \small
    \setlength{\tabcolsep}{4pt}
    \renewcommand{\arraystretch}{1.05}
    \begin{tabular}{lrcccc}
        \toprule
        \textbf{Chart type} & \textbf{n} & \textbf{Gemini} & \textbf{GPT} & \textbf{Llama} & \textbf{TinyChart} \\
        \midrule
        Grouped bar     & 248 & 95.18 & 67.91 & 48.39 & 54.23 \\
        Stacked bar     & 248 & 89.39 & 69.67 & 31.52 & 29.28 \\
        Line            & 252 & 87.68 & 38.53 & 28.10 & 18.32 \\
        Area            & 252 & 79.20 & 29.48 & 13.94 & 13.02 \\
        \midrule
        Overall         & 1000 & 87.83 & 51.26 & 30.41 & 28.61 \\
        \bottomrule
    \end{tabular}
    \caption{Single-pass RMS$_{\text{F1}}$ on WB-ChartExtract by chart type. Bar charts are easiest (fewer subsampled years, discrete heights); area charts are hardest (overlapping filled regions).}
    \label{tab:wb_by_type}
\end{table}

\subsection{Empirical Validation of Uncertainty Estimation}
\label{sec:uncertainty}
We assess whether ensemble disagreement predicts extraction quality by correlating each table-level uncertainty summary with final RMS$_{F1}$. On WB-ChartExtract, relative MAD is significantly anti-correlated with RMS$_{F1}$ (Spearman $\rho=-0.34$ for $U_{\text{med}}$, $\rho=-0.37$ for $U_{\text{mean}}$, and $\rho=-0.30$ for $U_{\text{max}}$; all $p<0.001$). Thus, higher ensemble disagreement reliably indicates lower extraction accuracy, making relative MAD a useful uncertainty estimate for downstream decisions.

\subsection{Convergence Detection}
\subsubsection{Convergence Detection Ablation}
\label{sec:convergence-detection-ablation}
We ablate convergence detection using Llama~4~Scout. With convergence detection \emph{off}, we always sample a fixed budget of $K_{\max}=20$ tables per image. With early stopping \emph{on}, we stop once the ensemble meets our convergence criterion, up to the same $K_{\max}$.

Table~\ref{tab:early_stopping_on_off} shows that early stopping achieves most of the ensemble gain over single-pass while using fewer samples on average. On WB-ChartExtract, fixed-budget ensembling improves RMS$_{\text{F1}}$ from 30.41 to 33.17 (+2.76), while early stopping reaches 33.14 (+2.73), retaining \textbf{99\%} of the ensembling boost with \textbf{1.24$\times$ fewer} samples (20.0$\rightarrow$16.11). The modest speedup on WB-ChartExtract reflects the benchmark's difficulty: with chart-type diversity and complex plots, the ensemble stabilizes slowly and early stopping triggers only for a subset of images (27.7\% convergence rate).

These results suggest convergence detection is practical for deployment, substantially reducing sampling cost while preserving most ensemble gains. Crucially, convergence is evaluated \emph{independently for each chart}: every image stops as soon as its own consecutive aggregates agree, so the method adaptively spends more samples on hard charts and fewer on easy ones, with no preset stopping point or per-dataset tuning required. Unless otherwise noted, we report results with early stopping on.

\begin{table}[t]
    \centering
    \small
    \setlength{\tabcolsep}{5pt}
    \renewcommand{\arraystretch}{1.05}
    \begin{tabular}{lcc}
        \toprule
        \textbf{Early stopping} & \textbf{RMS$_{\text{F1}}$} & \textbf{$\bar{S}$} \\
        \midrule
        Off (fixed budget)  & 33.17 & 20.00 \\
        On (early stopping) & 33.14 & 16.11 \\
        \bottomrule
    \end{tabular}
    \caption{Effect of early stopping on WB-ChartExtract using Llama~4~Scout. $\bar{S}$ is the average number of samples per image (lower is better).}
    \label{tab:early_stopping_on_off}
\end{table}

\subsubsection{Ensemble Size}
\label{sec:maximum-ensemble-size}

\begin{figure}[t]
    \centering
    \includegraphics[width=0.48\textwidth]{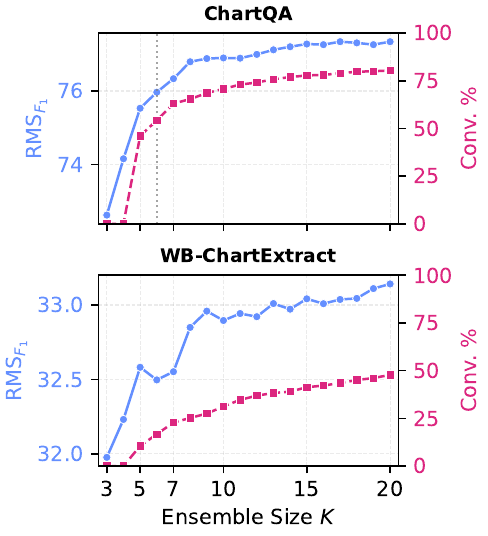}
    \caption{Effect of ensemble size $K$ on RMS$_{\text{F1}}$ and the percentage of examples that have met the convergence criterion by size $K$. Gray dashed lines mark the median stopping iteration under early stopping.}
    \label{fig:convergence_rate_comparison}
\end{figure}

Figure~\ref{fig:convergence_rate_comparison} shows how RMS$_{\text{F1}}$ and convergence evolve with ensemble size $K$ (Llama~4~Scout). On ChartQA ($T{=}2.0$), RMS$_{\text{F1}}$ rises sharply from $K{=}3$ to $K{=}7$ and largely saturates thereafter, with 80\% of images converged by $K{=}20$ and a median stopping iteration of $K{=}6$ (average samples per image $\bar{S}{=}9.34$). On WB-ChartExtract ($T{=}0.0$), gains are more gradual and convergence is slower: at $K{=}20$ only 27.7\% of images have converged, reflecting the benchmark's greater difficulty with diverse chart types. The median stopping iteration here is $K{=}20$ because fewer than half of the images converge before the cap, so the 50th-percentile image simply hits $K_{\max}$; as noted above, each chart still stops at its own convergence point, so this lower convergence rate reflects more genuinely hard charts rather than a uniform cutoff.

\subsection{Sampling Temperature}
\label{sampling-temperature}
Sampling temperature trades per-sample accuracy against ensemble diversity. Figure~\ref{fig:sampling-temperature} (Appendix~\ref{sec:temperature-figure}) plots $\mathrm{RMS}_{F1}$ vs.\ ensemble size for $T\in\{0.0,1.0,2.0\}$ using Llama~4~Scout: the optimum is benchmark-dependent. On ChartQA, $T{=}0.0$ wins for very small ensembles ($K{\le}4$) but $T{=}2.0$ takes over for $K{\ge}5$ (best ensemble 77.10) as added diversity outweighs reduced determinism; we therefore use $T{=}0.0$ for direct-prediction baselines and $T{=}2.0$ for self-ensembling. On WB-ChartExtract, $T{=}0.0$ is strictly best at every $K$ (best ensemble 33.14)---the harder charts cannot afford reduced per-sample accuracy---so we use $T{=}0.0$ for both. We tune temperature only on Llama~4~Scout and reuse the resulting choices ($T{=}0$ for direct-prediction baselines and on WB-ChartExtract self-ensembling; $T{=}2$ for ChartQA self-ensembling) across the other base models without per-model retuning, so reported gains for Qwen3-VL, Seed~1.6~Flash, DePlot, and TinyChart are not optimized for those models.

\section{Conclusion}
We introduced a model-agnostic self-ensembling method for chart data extraction that exploits VLM stochasticity across repeated runs to produce more accurate consensus tables than a single pass. The method also provides ensemble uncertainty estimates and a practical convergence-based early-stopping strategy. To better stress-test chart extraction, we presented WB-ChartExtract, a new benchmark derived from World Bank data with four chart types rendered by four plotting libraries, providing chart-type and cross-library diversity alongside higher data density. Across ChartQA and WB-ChartExtract, self-ensembling improves over single-pass VLM baselines in nearly every model-benchmark combination, with gains of up to 23\% relative.

\section{Limitations}
Our approach inherits several limitations from both the underlying VLM and our ensembling procedure. First, self-ensembling is bounded by base-model failure modes: it can only aggregate among the candidate tables the VLM produces. When the model makes systematic errors, aggregation may not correct them and can even reinforce consistent mistakes.

Second, our early-stopping criterion focuses on numeric stability and may miss ``semantic'' non-convergence. Because convergence is defined by relative changes in aggregated numeric cells, the ensemble may stabilize while the table structure remains incorrect, particularly when such structural mistakes are consistent across samples.

Third, our uncertainty estimate is heuristic and incomplete. Relative MAD provides a useful disagreement signal, but it is not a calibrated probability of correctness. In particular, uncertainty can be low even when the model is consistently wrong due to systematic errors, so it should be interpreted as a diagnostic rather than a formal confidence measure.

Fourth, WB-ChartExtract is synthetically rendered. While based on real World Bank time series and rendered across four plotting libraries (Matplotlib, Seaborn, Plotly, Bokeh) for stylistic diversity, the charts may not capture all real-world artifacts such as overlapping annotations, low-resolution scans, or non-standard layouts.

\bibliography{custom}

@software{WebPlotDigitizer,
    author = {Ankit Rohatgi},
    title = {WebPlotDigitizer},
    url = {https://automeris.io},
    version = {5.2},
}

@misc{chartqa,
      title={ChartQA: A Benchmark for Question Answering about Charts with Visual and Logical Reasoning}, 
      author={Ahmed Masry and Do Xuan Long and Jia Qing Tan and Shafiq Joty and Enamul Hoque},
      year={2022},
      eprint={2203.10244},
      archivePrefix={arXiv},
      primaryClass={cs.CL},
      url={https://arxiv.org/abs/2203.10244}, 
}

@misc{plotqa,
      title={PlotQA: Reasoning over Scientific Plots},
      author={Nitesh Methani and Pritha Ganguly and Mitesh M. Khapra and Pratyush Kumar},
      year={2020},
      eprint={1909.00997},
      archivePrefix={arXiv},
      primaryClass={cs.CV},
      url={https://arxiv.org/abs/1909.00997},
}

@misc{figureqa,
      title={FigureQA: An Annotated Figure Dataset for Visual Reasoning},
      author={Samira Ebrahimi Kahou and Vincent Michalski and Adam Atkinson and Akos Kadar and Adam Trischler and Yoshua Bengio},
      year={2017},
      eprint={1710.07300},
      archivePrefix={arXiv},
      primaryClass={cs.CV},
      url={https://arxiv.org/abs/1710.07300},
}

@misc{chartvlm,
      title={ChartX \& ChartVLM: A Versatile Benchmark and Foundation Model for Complicated Chart Reasoning}, 
      author={Renqiu Xia and Bo Zhang and Hancheng Ye and Xiangchao Yan and Qi Liu and Hongbin Zhou and Zijun Chen and Peng Ye and Min Dou and Botian Shi and Junchi Yan and Yu Qiao},
      year={2025},
      eprint={2402.12185},
      archivePrefix={arXiv},
      primaryClass={cs.CV},
      url={https://arxiv.org/abs/2402.12185}, 
}

@misc{onechart,
      title={OneChart: Purify the Chart Structural Extraction via One Auxiliary Token}, 
      author={Jinyue Chen and Lingyu Kong and Haoran Wei and Chenglong Liu and Zheng Ge and Liang Zhao and Jianjian Sun and Chunrui Han and Xiangyu Zhang},
      year={2024},
      eprint={2404.09987},
      archivePrefix={arXiv},
      primaryClass={cs.CV},
      url={https://arxiv.org/abs/2404.09987}, 
}

@inproceedings{tinychart,
    title = "{T}iny{C}hart: Efficient Chart Understanding with Program-of-Thoughts Learning and Visual Token Merging",
    author = "Zhang, Liang  and
      Hu, Anwen  and
      Xu, Haiyang  and
      Yan, Ming  and
      Xu, Yichen  and
      Jin, Qin  and
      Zhang, Ji  and
      Huang, Fei",
    editor = "Al-Onaizan, Yaser  and
      Bansal, Mohit  and
      Chen, Yun-Nung",
    booktitle = "Proceedings of the 2024 Conference on Empirical Methods in Natural Language Processing",
    month = nov,
    year = "2024",
    address = "Miami, Florida, USA",
    publisher = "Association for Computational Linguistics",
    url = "https://aclanthology.org/2024.emnlp-main.112/",
    doi = "10.18653/v1/2024.emnlp-main.112",
    pages = "1882--1898"
}

@misc{selfconsistency,
      title={Self-Consistency Improves Chain of Thought Reasoning in Language Models}, 
      author={Xuezhi Wang and Jason Wei and Dale Schuurmans and Quoc Le and Ed Chi and Sharan Narang and Aakanksha Chowdhery and Denny Zhou},
      year={2023},
      eprint={2203.11171},
      archivePrefix={arXiv},
      primaryClass={cs.CL},
      url={https://arxiv.org/abs/2203.11171}, 
}

@misc{ttc,
      title={Scaling LLM Test-Time Compute Optimally can be More Effective than Scaling Model Parameters}, 
      author={Charlie Snell and Jaehoon Lee and Kelvin Xu and Aviral Kumar},
      year={2024},
      eprint={2408.03314},
      archivePrefix={arXiv},
      primaryClass={cs.LG},
      url={https://arxiv.org/abs/2408.03314}, 
}

@inproceedings{revision,
author = {Savva, Manolis and Kong, Nicholas and Chhajta, Arti and Fei-Fei, Li and Agrawala, Maneesh and Heer, Jeffrey},
title = {ReVision: automated classification, analysis and redesign of chart images},
year = {2011},
isbn = {9781450307161},
publisher = {Association for Computing Machinery},
address = {New York, NY, USA},
url = {https://doi.org/10.1145/2047196.2047247},
doi = {10.1145/2047196.2047247},
booktitle = {Proceedings of the 24th Annual ACM Symposium on User Interface Software and Technology},
pages = {393–402},
numpages = {10},
location = {Santa Barbara, California, USA},
series = {UIST '11}
}

@INPROCEEDINGS {early,
author = { Leow, Wee Kheng and Huang, Weihua and Tan, Chew Lim },
booktitle = { Proceedings. Eighth International Conference on Document Analysis and Recognition },
title = {{ Associating Text and Graphics for Scientific Chart Understanding }},
year = {2005},
volume = {},
ISSN = {1520-5263},
pages = {580-584},
doi = {10.1109/ICDAR.2005.54},
url = {https://doi.ieeecomputersociety.org/10.1109/ICDAR.2005.54},
publisher = {IEEE Computer Society},
address = {Los Alamitos, CA, USA},
month =sep}

@inproceedings{deplot,
    title = "{D}e{P}lot: One-shot visual language reasoning by plot-to-table translation",
    author = "Liu, Fangyu  and
      Eisenschlos, Julian  and
      Piccinno, Francesco  and
      Krichene, Syrine  and
      Pang, Chenxi  and
      Lee, Kenton  and
      Joshi, Mandar  and
      Chen, Wenhu  and
      Collier, Nigel  and
      Altun, Yasemin",
    editor = "Rogers, Anna  and
      Boyd-Graber, Jordan  and
      Okazaki, Naoaki",
    booktitle = "Findings of the Association for Computational Linguistics: ACL 2023",
    month = jul,
    year = "2023",
    address = "Toronto, Canada",
    publisher = "Association for Computational Linguistics",
    url = "https://aclanthology.org/2023.findings-acl.660/",
    doi = "10.18653/v1/2023.findings-acl.660",
    pages = "10381--10399"
}

@inproceedings{early2,
author = {Mishchenko, Ales and Vassilieva, Natalia},
year = {2011},
month = {09},
pages = {115-120},
title = {Chart image understanding and numerical data extraction},
booktitle = {Proceedings of the 2011 Sixth International Conference on Digital Information Management (ICDIM)},
publisher = {IEEE},
doi = {10.1109/ICDIM.2011.6093320}
}

@inproceedings{early3,
author = {Al-Zaidy, Rabah A. and Giles, C. Lee},
title = {A machine learning approach for semantic structuring of scientific charts in scholarly documents},
year = {2017},
publisher = {AAAI Press},
booktitle = {Proceedings of the Thirty-First AAAI Conference on Artificial Intelligence},
pages = {4644–4649},
numpages = {6},
location = {San Francisco, California, USA},
series = {AAAI'17}
}

@misc{matcha,
      title={MatCha: Enhancing Visual Language Pretraining with Math Reasoning and Chart Derendering}, 
      author={Fangyu Liu and Francesco Piccinno and Syrine Krichene and Chenxi Pang and Kenton Lee and Mandar Joshi and Yasemin Altun and Nigel Collier and Julian Martin Eisenschlos},
      year={2023},
      eprint={2212.09662},
      archivePrefix={arXiv},
      primaryClass={cs.CL},
      url={https://arxiv.org/abs/2212.09662}, 
}

@misc{chartllama,
      title={ChartLlama: A Multimodal LLM for Chart Understanding and Generation}, 
      author={Yucheng Han and Chi Zhang and Xin Chen and Xu Yang and Zhibin Wang and Gang Yu and Bin Fu and Hanwang Zhang},
      year={2023},
      eprint={2311.16483},
      archivePrefix={arXiv},
      primaryClass={cs.CV},
      url={https://arxiv.org/abs/2311.16483}, 
}

@misc{chartast,
      title={ChartAssistant: A Universal Chart Multimodal Language Model via Chart-to-Table Pre-training and Multitask Instruction Tuning},
      author={Fanqing Meng and Wenqi Shao and Quanfeng Lu and Peng Gao and Kaipeng Zhang and Yu Qiao and Ping Luo},
      year={2024},
      eprint={2401.02384},
      archivePrefix={arXiv},
      primaryClass={cs.CV},
      url={https://arxiv.org/abs/2401.02384}, 
}

@inproceedings{plottwist,
    title = "Plot Twist: Multimodal Models Don{'}t Comprehend Simple Chart Details",
    author = "Razeghi, Yasaman  and
      Dasgupta, Ishita  and
      Liu, Fangyu  and
      Ramasesh, Vinay Venkatesh  and
      Singh, Sameer",
    editor = "Al-Onaizan, Yaser  and
      Bansal, Mohit  and
      Chen, Yun-Nung",
    booktitle = "Findings of the Association for Computational Linguistics: EMNLP 2024",
    month = nov,
    year = "2024",
    address = "Miami, Florida, USA",
    publisher = "Association for Computational Linguistics",
    url = "https://aclanthology.org/2024.findings-emnlp.342/",
    doi = "10.18653/v1/2024.findings-emnlp.342",
    pages = "5922--5937"
}

@misc{amplified,
      title={More Thinking, Less Seeing? Assessing Amplified Hallucination in Multimodal Reasoning Models}, 
      author={Chengzhi Liu and Zhongxing Xu and Qingyue Wei and Juncheng Wu and James Zou and Xin Eric Wang and Yuyin Zhou and Sheng Liu},
      year={2025},
      eprint={2505.21523},
      archivePrefix={arXiv},
      primaryClass={cs.CL},
      url={https://arxiv.org/abs/2505.21523}, 
}

@misc{spatialreasoninghard,
      title={Why Is Spatial Reasoning Hard for VLMs? An Attention Mechanism Perspective on Focus Areas}, 
      author={Shiqi Chen and Tongyao Zhu and Ruochen Zhou and Jinghan Zhang and Siyang Gao and Juan Carlos Niebles and Mor Geva and Junxian He and Jiajun Wu and Manling Li},
      year={2025},
      eprint={2503.01773},
      archivePrefix={arXiv},
      primaryClass={cs.CL},
      url={https://arxiv.org/abs/2503.01773}, 
}

@online{worldbank,
  author  = {{The World Bank}},
  title   = {World Development Indicators | DataBank: GDP growth (annual \%) (NY.GDP.MKTP.KD.ZG)},
  url     = {https://databank.worldbank.org/indicator/NY.GDP.MKTP.KD.ZG/1ff4a498/Popular-Indicators},
  urldate = {2025-12-12}
}

@misc{groq,
  author       = {{Groq, Inc.}},
  title        = {Groq is fast, low cost inference},
  year         = {2025},
  url          = {https://groq.com},
}

@Article{matplotlib,
  Author    = {Hunter, J. D.},
  Title     = {Matplotlib: A 2D graphics environment},
  Journal   = {Computing in Science \& Engineering},
  Volume    = {9},
  Number    = {3},
  Pages     = {90--95},
  publisher = {IEEE COMPUTER SOC},
  doi       = {10.1109/MCSE.2007.55},
  year      = 2007
}

@misc{anls,
      title={Scene Text Visual Question Answering}, 
      author={Ali Furkan Biten and Ruben Tito and Andres Mafla and Lluis Gomez and Marçal Rusiñol and Ernest Valveny and C. V. Jawahar and Dimosthenis Karatzas},
      year={2019},
      eprint={1905.13648},
      archivePrefix={arXiv},
      primaryClass={cs.CV},
      url={https://arxiv.org/abs/1905.13648}, 
}

@article{qwen,
      title={Qwen3-VL Technical Report}, 
      author={Shuai Bai and Yuxuan Cai and Ruizhe Chen and Keqin Chen and Xionghui Chen and Zesen Cheng and Lianghao Deng and Wei Ding and Chang Gao and Chunjiang Ge and Wenbin Ge and Zhifang Guo and Qidong Huang and Jie Huang and Fei Huang and Binyuan Hui and Shutong Jiang and Zhaohai Li and Mingsheng Li and Mei Li and Kaixin Li and Zicheng Lin and Junyang Lin and Xuejing Liu and Jiawei Liu and Chenglong Liu and Yang Liu and Dayiheng Liu and Shixuan Liu and Dunjie Lu and Ruilin Luo and Chenxu Lv and Rui Men and Lingchen Meng and Xuancheng Ren and Xingzhang Ren and Sibo Song and Yuchong Sun and Jun Tang and Jianhong Tu and Jianqiang Wan and Peng Wang and Pengfei Wang and Qiuyue Wang and Yuxuan Wang and Tianbao Xie and Yiheng Xu and Haiyang Xu and Jin Xu and Zhibo Yang and Mingkun Yang and Jianxin Yang and An Yang and Bowen Yu and Fei Zhang and Hang Zhang and Xi Zhang and Bo Zheng and Humen Zhong and Jingren Zhou and Fan Zhou and Jing Zhou and Yuanzhi Zhu and Ke Zhu},
	  journal={arXiv preprint arXiv:2511.21631},
      year={2025}
}

@misc{gpt,
  author       = {OpenAI},
  title        = {GPT-5.1: A smarter, more conversational ChatGPT},
  year         = {2025},
  month        = nov,
  howpublished = {\url{https://openai.com/index/gpt-5-1/}}
}

@misc{llama,
  author       = {{Meta AI}},
  title        = {The Llama 4 herd: The beginning of a new era of natively multimodal intelligence},
  year         = {2025},
  month        = apr,
  howpublished = {\url{https://ai.meta.com/blog/llama-4-multimodal-intelligence/}}
}

@misc{claude,
  author       = {Anthropic},
  title        = {Introducing Claude Opus 4.6},
  year         = {2026},
  howpublished = {\url{https://www.anthropic.com/news/claude-opus-4-6}}
}

@misc{gemini_flash_lite,
  author       = {Google},
  title        = {Gemini 3.1 Flash Lite},
  year         = {2026},
  howpublished = {\url{https://blog.google/innovation-and-ai/models-and-research/gemini-models/gemini-3-1-flash-lite/}},
  note         = {Accessed: 2026-05-04}
}

@misc{seed,
  author       = {{ByteDance Seed Team}},
  title        = {Introduction to Techniques Used in {Seed1.6}},
  year         = {2025},
  howpublished = {\url{https://seed.bytedance.com/en/blog/introduction-to-techniques-used-in-seed1-6}},
  note         = {Accessed: 2026-04-09}
}

@INPROCEEDINGS{deplot_cited,
  author={Biten, Ali Furkan and Tito, Rubèn and Mafla, Andrés and Gomez, Lluis and Rusiñol, Marçal and Mathew, Minesh and Jawahar, C.V. and Valveny, Ernest and Karatzas, Dimosthenis},
  booktitle={2019 International Conference on Document Analysis and Recognition (ICDAR)}, 
  title={ICDAR 2019 Competition on Scene Text Visual Question Answering}, 
  year={2019},
  volume={},
  number={},
  pages={1563-1570},
  doi={10.1109/ICDAR.2019.00251}}

@article{huber,
  title={Robust Estimation of a Location Parameter},
  author={Huber, Peter J.},
  journal={The Annals of Mathematical Statistics},
  volume={35},
  number={1},
  pages={73--101},
  year={1964},
  publisher={Institute of Mathematical Statistics}
}

@inproceedings{chartocr,
  title={{ChartOCR}: Data Extraction from Charts Images via a Deep Hybrid Framework},
  author={Luo, Junyu and Li, Zekun and Wang, Jinpeng and Lin, Chin-Yew},
  booktitle={Proceedings of the IEEE/CVF Winter Conference on Applications of Computer Vision (WACV)},
  pages={1917--1925},
  year={2021}
}

\appendix

\section{RMS Metric Details}
\label{sec:rms-details}
We evaluate chart data extraction using \emph{Relative Mapping Similarity} \citep[RMS]{deplot}, a table-matching metric that treats a table as an unordered set of mappings from a (row header, column header) pair to a cell value. We represent the predicted table as $P=\{p_i\}_{i=1}^{N}$ and the ground-truth table as $T=\{t_j\}_{j=1}^{M}$, where each entry is a triple $p_i=(p_i^{r},p_i^{c},p_i^{v})$ and $t_j=(t_j^{r},t_j^{c},t_j^{v})$. This set representation is invariant to row/column permutations.

RMS compares \emph{keys} (row/column headers) using a thresholded normalized Levenshtein distance $\mathrm{NL}_{\tau}(\cdot,\cdot)$, clipping distances above $\tau$ to 1. Keys are formed by concatenating headers: $k(p_i)=p_i^{r}\Vert p_i^{c}$ and $k(t_j)=t_j^{r}\Vert t_j^{c}$. Numeric values are compared by a clipped relative error
\[
D_{\theta}(p,t)=\min\Bigl(1,\frac{\lVert p-t\rVert}{\lVert t\rVert}\Bigr),
\]
and combined into an entry-level similarity
\[
s(p_i,t_j)=\Bigl(1-\mathrm{NL}_{\tau}(k(p_i),k(t_j))\Bigr)\Bigl(1-D_{\theta}(p_i^{v},t_j^{v})\Bigr),
\]
which is high only when both headers and values agree.

RMS then finds a minimum-cost bipartite matching between predicted and target entries (key-based), yielding an assignment matrix $X\in\mathbb{R}^{N\times M}$, and computes mapping-level precision and recall:
\[
\begin{aligned}
\mathrm{RMS}_{\text{prec}} &= \frac{1}{N}\sum_{i=1}^{N}\sum_{j=1}^{M} X_{ij}\, s(p_i,t_j), \\
\mathrm{RMS}_{\text{rec}}  &= \frac{1}{M}\sum_{i=1}^{N}\sum_{j=1}^{M} X_{ij}\, s(p_i,t_j).
\end{aligned}
\]
We report $\mathrm{RMS}_{F1}$, the harmonic mean of $\mathrm{RMS}_{\text{prec}}$ and $\mathrm{RMS}_{\text{rec}}$. Following \citet{deplot}, we set $\tau=0.5$ and $\theta=0.1$. Because predictions may be transposed, RMS also scores both the predicted table and its transpose and takes the higher $\mathrm{RMS}_{F1}$, yielding transposition invariance.

\section{ChartQA Chart-Type Breakdown}
\label{sec:chartqa-by-type}
Table~\ref{tab:chartqa_by_type} reports single-pass RMS$_{\text{F1}}$ on ChartQA broken down by chart type (classified using Gemini~3.1 Flash Lite \citep{gemini_flash_lite}). All models exhibit the same qualitative pattern: pie charts are by far the hardest category, followed by line charts and stacked bar charts, while vertical bar charts are easiest. TinyChart leads in every category, consistent with its ChartQA-specific fine-tuning providing a uniform advantage rather than benefiting only particular chart types. Pie charts stand out as especially challenging for the general-purpose models: Llama~4~Scout scores only 38.64, and even GPT-5.1 reaches just 51.43. These categories require inferring values from angles or slopes rather than aligned bar endpoints, explaining the larger gap between the specialist and generalist models.

\begin{table}[t]
    \centering
    \small
    \setlength{\tabcolsep}{4pt}
    \renewcommand{\arraystretch}{1.05}
    \begin{tabular}{lrccc}
        \toprule
        \textbf{Chart type} & \textbf{n} & \textbf{TinyChart} & \textbf{Llama~4~Scout} & \textbf{GPT-5.1} \\
        \midrule
        Vertical bar    & 673 & 99.16 & 83.71 & 93.53 \\
        Horizontal bar  & 380 & 97.35 & 76.09 & 87.13 \\
        Stacked bar     & 167 & 91.78 & 67.25 & 71.13 \\
        Line            & 211 & 86.59 & 65.73 & 75.91 \\
        Pie             &  78 & 81.32 & 38.64 & 51.43 \\
        \midrule
        Overall         & 1509 & 95.20 & 75.13 & 84.80 \\
        \bottomrule
    \end{tabular}
    \caption{Single-pass RMS$_{\text{F1}}$ on ChartQA by chart type. Pie and line charts are the hardest categories across all models; the specialist TinyChart leads uniformly.}
    \label{tab:chartqa_by_type}
\end{table}

\section{Aggregation Method Ablation}
\label{sec:aggregation-method}
We ablate the cell-wise aggregation function using Llama~4~Scout on WB-ChartExtract, comparing five strategies (Table~\ref{tab:aggregation_ablation}): mean, Huber estimator \cite{huber}, median, \emph{Weighted Confidence} (retaining the top 60\% most consistent extractions based on similarity to the ensemble median), and \emph{RANSAC-style} (removing outliers beyond 2$\times$MAD from the median). Median performs best so we use it in all experiments.

\begin{table}[t]
    \centering
    \small
    \begin{tabular}{lc}
        \toprule
        \textbf{Aggregator} & \textbf{WB-ChartExtract} \\
        \midrule
        Mean & 30.49 \\
        RANSAC & 31.18 \\
        Weighted Confidence & 31.41 \\
        Huber & 32.82 \\
        \textbf{Median} & \textbf{33.14} \\
        \bottomrule
    \end{tabular}
    \caption{Aggregation method ablation using Llama~4~Scout on WB-ChartExtract (RMS$_{\text{F1}}$).}
    \label{tab:aggregation_ablation}
\end{table}

\section{Sensitivity to Convergence Hyperparameters}
\label{sec:sensitivity}

Our early-stopping rule declares convergence when, between two consecutive ensemble updates, at least a fraction \emph{coverage} of aggregated cells change by no more than a relative \emph{tolerance}, and this holds for \emph{patience} consecutive updates. Increasing patience or coverage, or decreasing tolerance, makes the criterion stricter, typically requiring more samples but yielding marginal accuracy gains. The hyperparameter ablations in this section, as well as those for early stopping (Section~\ref{sec:convergence-detection-ablation}), aggregation (Appendix~\ref{sec:aggregation-method}), and cluster pruning (Appendix~\ref{sec:cluster-pruning-threshold}), are conducted with Llama~4~Scout on WB-ChartExtract, and the resulting defaults are then applied uniformly to all other base models without per-model retuning.

Tables~\ref{tab:sens_patience}--\ref{tab:sens_tolerance} examine this trade-off on WB-ChartExtract using Llama~4~Scout. \textbf{Patience} has the largest compute swing (Table~\ref{tab:sens_patience}): increasing it from 1 to 3 raises $\bar{S}$ from 12.51 to 18.70, while RMS$_{\text{F1}}$ rises only marginally (32.88$\rightarrow$33.14$\rightarrow$33.23) with clearly diminishing gains beyond patience=2. We use patience=2 as a default because it captures most of the accuracy gain over patience=1 (+0.26\,pp) while avoiding the disproportionate compute cost of patience=3 (+2.59 samples for only +0.09\,pp). Varying \textbf{coverage} (90\%$\rightarrow$97.5\%) modestly increases $\bar{S}$ with negligible change in RMS$_{\text{F1}}$ ($\le$0.06\,pp; Table~\ref{tab:sens_coverage}). \textbf{Tolerance} offers a similar cost--accuracy trade-off (Table~\ref{tab:sens_tolerance}): tighter tolerance (0.1\%) increases $\bar{S}$ with negligible improvement in RMS$_{\text{F1}}$ (+0.05\,pp), while looser tolerance (10\%) stops much earlier ($\bar{S}$ drops 25\%) at a small cost in RMS$_{\text{F1}}$ (0.24\,pp).

These hyperparameters therefore tune the cost--accuracy frontier. Unless otherwise noted, we use patience$=2$, coverage$=95\%$, and tolerance$=1\%$.

\begin{table}[t]
    \centering
    \small
    \setlength{\tabcolsep}{5pt}
    \begin{tabular}{lcc}
        \toprule
        \textbf{Patience} & \textbf{RMS$_{\text{F1}}$} & \textbf{$\bar{S}$} \\
        \midrule
        1           & 32.88 & 12.51 \\
        2 (default) & 33.14 & 16.11 \\
        3           & 33.23 & 18.70 \\
        \bottomrule
    \end{tabular}
    \caption{Sensitivity to \emph{patience}, fixing coverage=$95\%$ and tolerance=$1\%$. $\bar{S}$ is the average samples per image. Llama~4~Scout on WB-ChartExtract.}
    \label{tab:sens_patience}
\end{table}

\begin{table}[t]
    \centering
    \small
    \setlength{\tabcolsep}{5pt}
    \begin{tabular}{lcc}
        \toprule
        \textbf{Coverage} & \textbf{RMS$_{\text{F1}}$} & \textbf{$\bar{S}$} \\
        \midrule
        90\%           & 33.18 & 15.24 \\
        95\% (default) & 33.14 & 16.11 \\
        97.5\%         & 33.20 & 16.58 \\
        \bottomrule
    \end{tabular}
    \caption{Sensitivity to \emph{coverage}, fixing patience=$2$ and tolerance=$1\%$. Llama~4~Scout on WB-ChartExtract.}
    \label{tab:sens_coverage}
\end{table}

\begin{table}[t]
    \centering
    \small
    \setlength{\tabcolsep}{5pt}
    \begin{tabular}{lcc}
        \toprule
        \textbf{Tolerance} & \textbf{RMS$_{\text{F1}}$} & \textbf{$\bar{S}$} \\
        \midrule
        0.1\%          & 33.19 & 16.70 \\
        1\% (default)  & 33.14 & 16.11 \\
        10\%           & 32.90 & 12.11 \\
        \bottomrule
    \end{tabular}
    \caption{Sensitivity to \emph{tolerance}, fixing patience=$2$ and coverage=$95\%$. Llama~4~Scout on WB-ChartExtract.}
    \label{tab:sens_tolerance}
\end{table}

\section{Cluster Pruning Threshold}
\label{sec:cluster-pruning-threshold}
We ablate the cluster pruning threshold, which sets the minimum fraction of sampled tables in which a row/column label must appear to be retained after clustering. Higher thresholds remove low-support (often spurious) labels, while lower thresholds retain more labels but risk noise. Using Llama~4~Scout on WB-ChartExtract, we test thresholds in $\{0.0,0.1,0.2,0.3,0.4,0.5\}$, where $0.0$ denotes no pruning.

Table~\ref{tab:pruning_threshold} reports RMS$_{\text{F1}}$. The best threshold is 0.2 (33.14), with 0.1--0.3 forming a plateau within 0.3\,pp; performance degrades on both sides, with no pruning ($0.0$) at 32.00 and aggressive pruning ($0.5$) at 31.89. We use 0.2 as the default and apply this single value across all base models rather than tuning per model, since per-model tuning would risk cherry-picking and complicate cross-model comparisons. The optimum may differ for Qwen3-VL, Seed~1.6~Flash, DePlot, and TinyChart, which we did not separately ablate; for deployment, we recommend any value in the 0.1--0.3 range.

\begin{table}[t]
    \centering
    \small
    \begin{tabular}{lc}
        \toprule
        \textbf{Pruning Threshold} & \textbf{WB-ChartExtract} \\
        \midrule
        0.0 (no pruning) & 32.00 \\
        0.1              & 32.84 \\
        0.2 (default)    & \textbf{33.14} \\
        0.3              & 32.93 \\
        0.4              & 32.27 \\
        0.5              & 31.89 \\
        \bottomrule
    \end{tabular}
    \caption{Effect of the cluster pruning threshold on RMS$_{\text{F1}}$ using Llama~4~Scout on WB-ChartExtract.}
    \label{tab:pruning_threshold}
\end{table}

\section{Additional Metrics: RNSS and RD}
\label{sec:rnss-rd}
RMS$_{\text{F1}}$ jointly scores headers and values through a key-aware matching. To check that our gains are not an artifact of this particular metric, we additionally report two value-centric, header-agnostic metrics on WB-ChartExtract.

\paragraph{RNSS.} Relative Number Set Similarity \citep{chartocr} treats each table as a multiset of numeric values (headers and indices excluded). Given predicted values $\{p_i\}_{i=1}^{N}$ and ground-truth values $\{g_j\}_{j=1}^{M}$, we form the relative-distance matrix $d_{ij}=\min(1,|p_i-g_j|/|g_j|)$ (with $g_j{=}0$ scored as $0$ iff $p_i{=}0$, else $1$), pad the smaller set with dummy entries of distance $1$, and compute a minimum-cost bipartite matching. Then
\[
\mathrm{RNSS}=1-\frac{\sum_{\text{matched}} d_{ij} + |N-M|}{\max(N,M)},
\]
which penalizes both inaccurate values and cardinality mismatches (higher is better).

\paragraph{RD.} Relative Deviation isolates numeric closeness from cardinality: it is the mean of $d_{ij}$ over the matched real value pairs (lower is better), capturing how close matched numbers are irrespective of how many were matched.

Table~\ref{tab:rnss_rd} reports both. \textbf{RNSS} improves with self-ensembling for every base model (from $+0.5$ for TinyChart and DePlot up to $+4.7$ for Llama~4~Scout), mirroring the RMS$_{\text{F1}}$ trend and confirming that ensembling recovers more correct numbers. \textbf{RD} is more mixed: it decreases (improves) for four of the five base models---TinyChart, DePlot, Qwen3-VL, and Seed~1.6~Flash---but rises slightly for Llama~4~Scout ($19.96\rightarrow20.70$). This is consistent with the error analysis (\S\ref{sec:error_analysis}): Llama~4~Scout also posts the largest RNSS gain ($+4.7$), so ensembling recovers many additional matched datapoints, but the newly drawn-in matches are lower-precision and dilute per-match closeness, nudging RD upward.

\begin{table}[t]
    \centering
    \small
    \setlength{\tabcolsep}{5pt}
    \renewcommand{\arraystretch}{1.1}
    \begin{tabular}{lcccc}
        \toprule
        & \multicolumn{2}{c}{\textbf{RNSS} ($\uparrow$)} & \multicolumn{2}{c}{\textbf{RD} ($\downarrow$)} \\
        \cmidrule(lr){2-3}\cmidrule(lr){4-5}
        \textbf{Model} & Single & +Self-ens. & Single & +Self-ens. \\
        \midrule
        TinyChart      & 36.71 & 37.25 & 22.26 & 21.48 \\
        DePlot         & 30.95   & 31.48   & 26.86   & 25.93 \\
        Qwen3-VL       & 62.56 & 64.38 & 10.57 & 10.38 \\
        Seed~1.6~Flash & 56.34   & 59.94   & 17.59   & 16.18 \\
        Llama~4~Scout  & 51.28 & 55.95 & 19.96 & 20.70 \\
        \bottomrule
    \end{tabular}
    \caption{RNSS and RD on WB-ChartExtract, single-pass vs.\ self-ensembled. RNSS (higher is better) improves for every model; RD (lower is better) improves for all base models except Llama~4~Scout, whose RD rises slightly.}
    \label{tab:rnss_rd}
\end{table}

\section{Accuracy--Cost Tradeoff}
\label{sec:cost-figure}
\begin{figure*}[t]
    \centering
    \includegraphics[width=\textwidth]{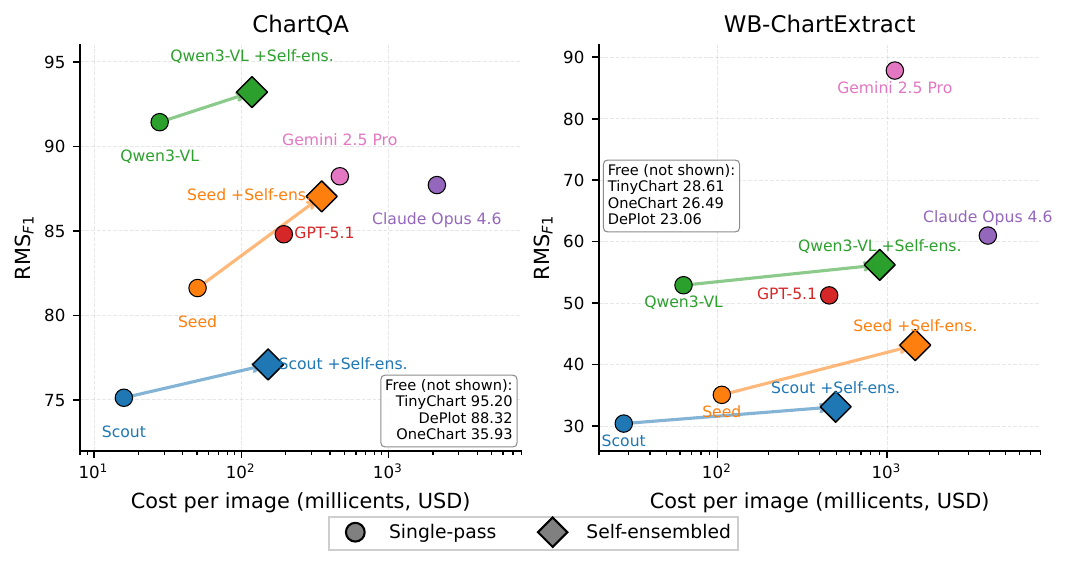}
    \caption{Accuracy--cost tradeoff on ChartQA (left) and WB-ChartExtract (right). Each base model's single-pass result (circle) is connected by an arrow to its self-ensembled variant (diamond); the $x$-axis is per-image API cost on a log scale. Self-ensembling consistently increases RMS$_{\text{F1}}$ at a higher per-image cost. TinyChart, OneChart, and DePlot run locally at zero API cost and are annotated separately, as they cannot be placed on the log-cost axis.}
    \label{fig:cost_efficiency}
\end{figure*}

Figure~\ref{fig:cost_efficiency} visualizes the accuracy--cost tradeoff discussed in \S\ref{sec:experiments}: it places each model's single-pass and self-ensembled RMS$_{\text{F1}}$ (Table~\ref{tab:main_results}) against its per-image API cost (Table~\ref{tab:cost}) on a log scale, making the upward-and-rightward shift induced by self-ensembling explicit.

\section{Sampling Temperature Ablation}
\label{sec:temperature-figure}
\begin{figure}[t]
    \centering
    \includegraphics[width=0.48\textwidth]{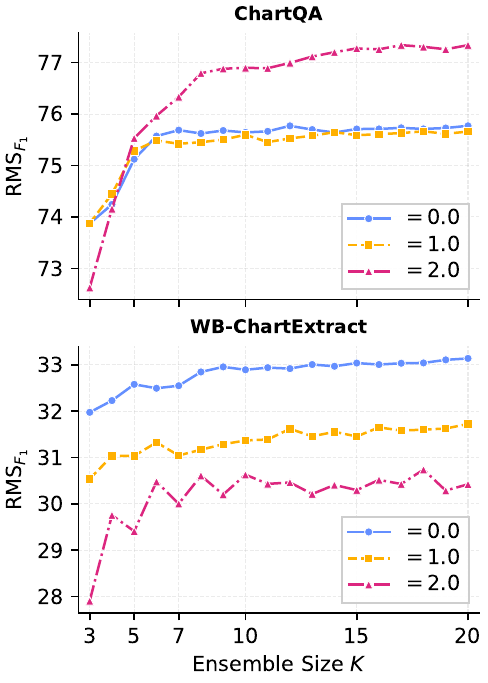}
    \caption{Effect of sampling temperature on ensemble performance using Llama~4~Scout on ChartQA (top) and WB-ChartExtract (bottom).}
    \label{fig:sampling-temperature}
\end{figure}

Figure~\ref{fig:sampling-temperature} reports the sampling-temperature ablation discussed in \S\ref{sampling-temperature}, plotting RMS$_{\text{F1}}$ against ensemble size for $T\in\{0.0,1.0,2.0\}$ on both benchmarks.

\end{document}